\documentclass[letterpaper, 10 pt, conference]{ieeeconf}  % Comment this line out if you need a4paper

\IEEEoverridecommandlockouts                              % This command is only needed if 
\usepackage{amsmath}
\usepackage{amsthm}
\usepackage{mathtools}
\usepackage{amssymb}
\usepackage{booktabs}
\usepackage{array}
\usepackage{multirow}
\usepackage{rotating}
\usepackage{hyperref}
\usepackage{cleveref}
\theoremstyle{definition}
\newtheorem{definition}{Definition}
\crefname{Assumption}{assumption}{assumptions}
\crefname{Definition}{definition}{definitions}

\usepackage[dvipsnames]{xcolor}
\definecolor{tabblue}{RGB}{31, 119, 180}
\definecolor{taborange}{RGB}{255, 127, 14}
\definecolor{tabgreen}{RGB}{44, 160, 44}

\title{\LARGE \bf
Problem Space Transformations for Out-of-Distribution \\ Generalisation in Behavioural Cloning
}

\author{Kiran Doshi$^{1}$, Marco Bagatella$^{1}$ and Stelian Coros$^{1}$% <-this % stops a space
\thanks{$^{1}$All authors are part of the Department of Computer Science at ETH Zürich, Zürich, Switzerland. All correspondence to
        {\tt\small doshik@ethz.ch}.}%
}

\begin{document}

\maketitle
\thispagestyle{empty}
\pagestyle{empty}

%%%%%%%%%%%%%%%%%%%%%%%%%%%%%%%%%%%%%%%%%%%%%%%%%%%%%%%%%%%%%%%%%%%%%%%%%%%%%%%%
\begin{abstract}

The combination of behavioural cloning and neural networks has driven significant progress in robotic manipulation.
    As these algorithms may require a large number of demonstrations for each task of interest, they remain fundamentally inefficient in complex scenarios, in which finite datasets can hardly cover the state space.
    One of the remaining challenges is thus out-of-distribution (OOD) generalisation, i.e. the ability to predict correct actions for states with a low likelihood with respect to the state occupancy induced by the dataset.
    This issue is aggravated when the system to control is treated as a black-box, ignoring its physical properties.
    This work highlights widespread properties of robotic manipulation, specifically pose equivariance and locality.
    We investigate the effect of the choice of problem space on OOD performance of BC policies and how transformations arising from characteristic properties of manipulation can be employed for its improvement.
    Through controlled, simulated and real-world experiments, we empirically demonstrate that these transformations allow behaviour cloning policies, using either standard MLP-based one-step action prediction or diffusion-based action-sequence prediction, to generalise better to certain OOD problem instances. Code is available at \url{https://github.com/kirandoshi/pst_ood_gen}.

\end{abstract}
% \vspace{3mm}

%%%%%%%%%%%%%%%%%%%%%%%%%%%%%%%%%%%%%%%%%%%%%%%%%%%%%%%%%%%%%%%%%%%%%%%%%%%%%%%%
\section{INTRODUCTION}

The behavioural cloning (BC) paradigm \cite{bain1995framework, ross2010efficient} has been the foundation of recent advances in robotic manipulation \cite{chi2023diffusionpolicy, zhao2024aloha, o2024open}.  
BC is particularly promising for robot manipulation, as humans are very proficient in general manipulation, and can quickly learn to collect demonstrations when given a well-designed interface \cite{zhao2023learning}.
An important benefit of using this data to train a robot policy is that it can be collected on the real system, thus avoiding the sim-to-real gap.
However, as a supervised learning method, BC requires the collected data to cover the workspace
with relatively high density \cite{pmlr-v15-ross11a, laskey2017dart, belkhale2024data}.
Neural networks trained with BC, and more generally functions estimated through supervised learning, hardly generalise outside the support of the training data, i.e. "out-of-distribution" (OOD)~\cite{zhang2021understanding, liu2021towards}.
% Policy prediction for OOD states can be arbitrary, which poses a safety risk.
Avoiding OOD states by providing sufficient data coverage can quickly become infeasible.
This is particularly aggravating for robotic manipulation, as collection of human demonstrations remains time intensive and thus expensive \cite{zhao2024aloha}. \looseness -1

% This work highlights and leverages practical assumptions on object-centric manipulation tasks, and explores a family of problem space transformations that enable OOD generalisation with respect to the original problem space.
% We observe that these transformations are a crucial design component for learning-based control of manipulators, and enable policies learned through simple BC to perform well on OOD states.

The aim of this work is to investigate the effect of the choice of the problem space on the ability of a BC policy to generalize OOD.
The focus is on robotic manipulation tasks with rigid bodies and low-dimensional state information.
Therefore, practical assumptions on object-centric manipulation tasks are highlighted and leveraged to explore a family of associated problem space transformations.
We observe that these transformations are a crucial design component for learning-based control of manipulators, and enable policies learned through BC to perform well on OOD states.
Our focus is not that of solving OOD generalisation as a whole, but rather to carefully evaluate how careful transformation of the problem space in a common setup impacts generalization.

We thus present three main contributions:
\begin{itemize}
    \item we formalise the OOD problem in the context of manipulation;
    \item we describe several transformations of the problem space that embed these properties;
    \item we provide experimental results demonstrating that the choice of problem space transformation significantly impacts the ability of OOD generalisation for three simulated robotic manipulation tasks and a real-world pushing task.
\end{itemize}

\begin{figure*}[h!]
    \centering
    \vspace{-7mm}
    \includegraphics[width=0.9\linewidth]{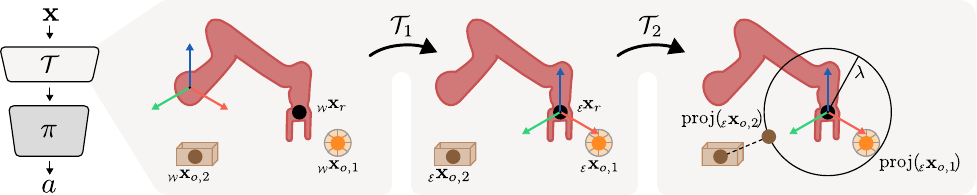}
    \vspace{-3mm}
    \caption{\looseness -1 Overview of the proposed transformations of the behaviour cloning problem space. The base problem space is with a (arbitrary, e.g. at the base of the robot) fixed Cartesian coordinate system $\mathcal{W}$ in which the pose of the end-effector as well as the objects are measured. $\mathcal{T}_1$ transforms the problem space such that the all poses are measured with respect to the moving Cartesian frame of the end-effector $\mathcal{E}$. $\mathcal{T}_2$ projects all values to a $\lambda$-ball centred at the origin of $\mathcal{E}$.}
    \vspace{-4mm}
    \label{fig:transformations_overview}
\end{figure*}

After introducing related works and preliminaries in Sections \ref{sec:related} and \ref{sec:preliminaries} respectively, we present our main contributions in Section \ref{sec:problem_space_transformation} and validate it empirically in Section \ref{sec:result}. An overview of the proposed problem space transformations can be seen in \Cref{fig:transformations_overview}.

\section{Related Work}

\label{sec:related}

\paragraph{Imitation Learning and Behavioural Cloning}

The core idea of imitation learning is that of extracting a control policy from high-quality data \cite{schaal1999imitation, osa2018algorithmic}.
Imitating expert trajectories can be framed in various formulations, encompassing both offline \cite{ross2010efficient, kim2022demodice} and online \cite{ho2016generative, garg2021iq} methods.
A core technique that has recently risen to prominence is that of behavioural cloning (BC) \cite{bain1995framework, ross2010efficient}, which casts the problem as supervised learning, and optimizes a policy by minimizing the distance of its output to expert actions.
Due to its offline nature, BC notoriously suffers from accumulating errors \cite{ross2010efficient}.
However, BC has been instrumental to multiple recent advances in robotic manipulation \cite{zhao2023learning, zhao2024aloha}, in which good demonstrations can be easily collected \cite{kumar2022should}.
In this context, diffusion-based parametrizations \cite{chi2023diffusionpolicy} have arisen as an expressive policy class, capable of modelling multi-modal behaviours.
Finally, formal understanding of BC has also gradually advanced \cite{block2023butterfly, foster2024behavior}, and performance guarantees have become more practical \cite{xu2020error, block2024provable}.

\paragraph{Using Robotic Manipulation Properties in Policy Learning}
Past work has explored exploiting different invariances and equivariances in robot manipulation learning.
In \cite{simeonov2022neural}, learnt $\mathrm{SE}(3)$-equivariant object representations, which have an object point cloud as input, are deployed to enable a pick and place system that requires only a few demonstrations.
In \cite{sim3_equivariant_policies}, $\mathrm{SIM}(3)$ equivariance (which additionally to $\mathrm{SE}(3)$ includes scale equivariance) is embedded in the object representation learning module (again from point clouds) as well as in the BC policy network module.
The work in \cite{yang2024equibot} takes a similar approach, though adding a $\mathrm{SIM}(3)$-equivariant Diffusion Policy as the BC policy module.
Recent work additionally explores the usefulness of the \textit{locality} of manipulation problems to increase sample efficiency by predicting actions as displacements to points in the scene point cloud \cite{zhang2024leveraging_locality} (thus assuming 3D sensing) or by leveraging gaze data \cite{takizawa2025enhancing}.
% All of the above works assume that the scene entities are sensed as point clouds.
Some past work also looks at the benefits of introducing 
$\mathrm{SO}(2)$-equivariance to online \cite{wang2022mathrmso2equivariantreinforcementlearning} and offline RL \cite{tangri2024equivariant_offline_RL} where the scene entity poses are assumed to be available.

\paragraph{Out-of-Distribution Generalisation}
% Chat GPT generated in the comments
% Out-of-Distribution (OOD) generalization is a fundamental challenge in machine learning, where models trained on a given distribution must generalize effectively to unseen distributions. Various approaches have been proposed to address this problem, particularly through the lens of invariance learning and risk minimization.
A large interest was taken in the problem of OOD generalisation (sometimes also called domain generalisation) in the area of image classification, where pure empirical risk minimisation (ERM) would, for instance, produce classifiers relying on the background instead of the subject of an image \cite{beery2018recognition}.
A prominent paradigm is to view the problem as a minimization of the worst-case error over a set of possible environments, with the aim to perform well across all of them \cite{arjovsky2019invariant}.
This approach, called Invariant Risk Minimization (IRM), aims to find a predictor which balances prediction accuracy with invariance across different environments.
Further work such as \cite{krueger2021out} and \cite{ahuja2021invariant} have built upon IRM. 
An alternative line of work views the data as a composition of semantic and non-semantic components \cite{ahmed2021systematic, fu2021learning}. In this case, the learned predictors should perform well when there is a covariate shift in the non-semantic distribution. 
For this purpose, \cite{ahmed2021systematic} assume the existence of minority and majority groups in the data, and introduce a loss term encouraging similar predictive distributions over both groups.
A related work in the context of sequential decision making \cite{fu2021learning} disentangles semantic components in images by assuming access to a reward function that is informative of the nature of the task.
These works assume that the dataset includes additional structure which provides implicit direction for a method to determine robust features.
Our works aims to solve a related, but different problem: the objective is known (i.e., extending the range of robotic tasks), but no additional information is available except for state-action trajectories (see \ref{sec:problem_space_transformation}).
We thus propose to leverage general inductive biases specific to physical systems, rather than instance-specific information which is, in our case, not available.

\section{Preliminaries}
\label{sec:preliminaries}

\subsection{Behavioural Cloning}
We assume that the data is collected in a finite-horizon Markov Decision Process (MDP) modelled as tuple $\mathcal{M} = (\mathcal{X}, \mathcal{A}, P,  \mathcal{R}, \mu_0, H)$, 
where $\mathcal{X}$ is the state space, $\mathcal{A}$ is the action space, $\mathcal{R}: \mathcal{X \times A} \to \mathbb{R}$ is the reward function, 
$P: \mathcal{X \times A} \to \Delta(\mathcal{X})$ is the dynamics transition probability function, $\mu_0 \in \Delta(\mathcal{X})$ is the initial state distribution and $H$ is the horizon. 
BC learns a stationary parameterised policy $\pi_\theta: \mathcal{X}^{T_X} \to \mathcal{A}^{T_A}$ from a dataset of $K$ task rollouts $\mathcal{D} = \{\tau_1, ..., \tau_K\}$ with $\tau_k = \{(x_0, a_0), ..., (x_H, a_H)\}$, $x_0 \sim \mu_0$, $x_{t+1} \sim P(x_t, a_t)$ and $a_t \sim \pi_d(s_t)$. In this case $\pi_d$ represents the demonstrator's policy. 
In the general case, the policy predicts a sequence of actions with $T_A$ steps based on an state sequence with $T_X$ steps.
We utilise two different methods to train BC polices in this work. 
(i) The first assumes that $T_X = T_A = 1$.
In that case, the policy $\pi_\theta$ is a deterministic multi-layer perceptron (MLP) and is trained by regressing actions with a loss function $\mathcal{L} = \sum_{(x, a) \in \mathcal{D}} D(a, \pi_{\theta}(x))$, where $D$ is an appropriate distance metric.
(ii) In the second case we allow both to be larger than 1, i.e. $T_X >1, ~T_A >1$. 
In this case the policy is trained as a Denoising Diffusion Probabilistic Model, which is known as a \textit{Diffusion Policy}~\cite{chi2023diffusionpolicy}.
In both cases the loss constitutes a \textit{proxy} objective, as the actual goal is to maximise the policy's returns: $\mathop{\mathbb{E}}_{\mu_0, \pi_\theta, P} \sum_{t=0}^{H-1} R(x_t, a_t)$.

\subsection{Robotic Manipulation and Problem Space}
In this work, we consider robotic manipulation tasks.
We define the space of state-action tuples as the problem space $\mathcal{P} = \mathcal{X} \times \mathcal{A}$.
As multiple MDPs can model the same environment, the problem space is generally chosen by the designer of the system. 
A common choice of state and action space by practitioners is as follows \cite{robot_learning_review_konidaris}.
The state space will include proprioceptive information $\mathbf{x}_{r}$, e.g. in the form of joint positions or the end-effector (EE) pose.
Furthermore, the poses of the entities in the scene would be included, resulting in $\mathbf{x} = [\mathbf{x}_{r}, (\mathbf{x}_{o,i})_1^{N_0}]$, where $N_O$ is the number of entities.
The action space is usually a set point for the low level robot control, either at joint or at EE level $\mathbf{a} = [\mathbf{a}_r]$.
As forward and inverse kinematics (calculation of EE pose from joint position values and its inverse) are usually accessible, we assume that both action and proprioception are expressed as EE poses without loss of generality. 
We also assume that the action is given as an offset to the current EE position, though it is also common to include it as the next EE position or as a velocity. 
We leave out the state and change of the gripper in state and action space respectively for the sake of brevity.
While we focus on this specific setting due to its prominence in learning-based robotics, we remark that several other problem spaces are common, each of which which would induce different properties and transformations.

\subsection{Out of Distribution Generalisation and BC}
Out-of-distribution (OOD) generalisation is the desirable capability of a model to return reasonable predictions for unseen data points.
We provide a practical, more specific definition in the context of this work.
As the learned model $\pi_\theta$ operates over states, we introduce a state occupancy $\Omega \in \Delta(\mathcal{S})$ such that samples in the dataset $\mathcal{D}$ can be considered to be drawn independently and identically distributed (iid) from it.
For a given choice of $\Omega$ (e.g. the solution of MLE in a class of smooth densities) and a threshold $\epsilon$, in-distribution generalisation occurs when the learned policy $\pi_\theta$ returns the unseen demonstrator's action for a state $\mathbf{x} \not \in \mathcal{D}$, but with $\Omega(\mathbf{x}) \geq \epsilon$.
We can thus introduce an in-distribution manifold $\mathcal{\hat X} = \{\mathbf{x} \in \mathcal{X} \mid \Omega(\mathbf{x}) \geq \epsilon\}$.
Similarly, we define

\begin{definition}[OOD generalisation, informal]
\label{def:ood}
    A policy $\pi$ is capable of OOD generalisation if its error is low for arbitrary states $\mathbf{x} \not \in \mathcal{D}$ such that $\Omega(\mathbf{x}) \leq \epsilon$.
    
\end{definition}
Let us consider a desired manifold of states $ \mathcal{X^\star} \supset \mathcal{\hat X}$ including OOD data points.
In general, a policy trained through BC on $\mathcal{D}$ will not generalise to $\mathcal{X^\star}$, as supervised learning assumes that the training and test data is iid. 
Further assumptions on the problem space are thus needed to enable OOD generalisation. 

\begin{figure}
    \centering
    \includegraphics[width=1.0\linewidth]{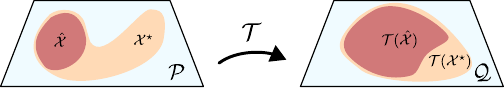}
    \caption{Visualisation of the effect of the proposed transformation on the in-distribution manifold $\hat{\mathcal{X}}$ and on the desired manifold $\mathcal{X}^\star$. The aim is that in the transformed space both manifolds are aligned such that 
    %the policy which achieves high reward in $\mathcal{T}(\hat{\mathcal{X}})$ also does so in $\mathcal{T}(\mathcal{X}^{\star})$}.
    OOD states in $\mathcal{P}$ receive supervision signal in $\mathcal{Q}$.}
    \vspace{-4mm}
    \label{fig:minimisation_visualisation}
\end{figure}

% This figure belongs to the results section but place here already for better distribution.
\begin{figure*}[h]
        \centering
        \vspace{-6mm}
        \begin{turn}{90}
        \hspace{-4mm}
        MLP
        \end{turn}
        \begin{minipage}{0.97\textwidth}
            \centering
            \includegraphics[height=31mm]{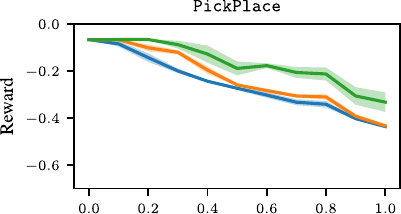}
            \includegraphics[height=31mm]{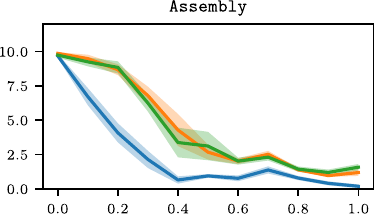}
            \includegraphics[height=31mm]{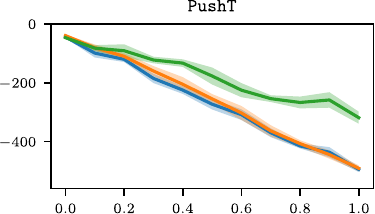}
        \end{minipage}        
        
        \vspace{2mm}

        \rule{\textwidth}{0.6pt}

        \vspace{2mm}

        \begin{turn}{90}
        \hspace{-10mm}
        Diffusion Policy
        \end{turn}
        \begin{minipage}{0.97\textwidth}
            \centering
            \includegraphics[height=31.16mm]{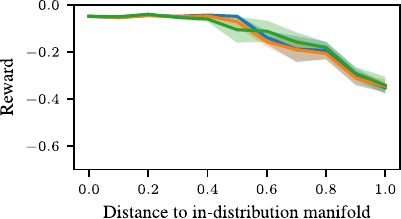}
            \includegraphics[height=31.16mm]{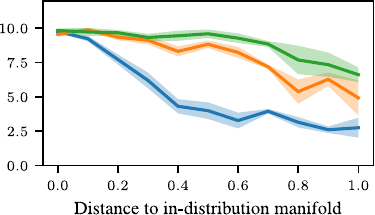}
            \includegraphics[height=31.16mm]{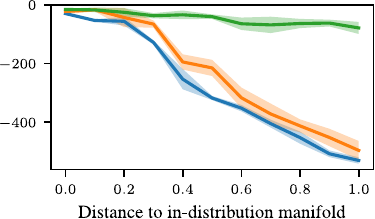}
        \end{minipage}

        \vspace{0.9em}
        \centering
        \raisebox{0.15em}{\textcolor{tabblue}{\rule{1em}{0.2em}}} Baseline \quad \raisebox{0.15em}{\textcolor{taborange}{\rule{1em}{0.2em}}} $\mathcal{T}_1$ \quad \raisebox{0.15em}{\textcolor{tabgreen}{\rule{1em}{0.2em}}} $\mathcal{T}_2$
        \vspace{-0.4em}
        \caption{\looseness -1 Comparison of BC policies trained in the original problem space $\mathcal{P}$, in $\mathcal{T}_1(\mathcal{P})$ and $\mathcal{T}_2(\mathcal{P})$ on in-distribution and OOD initial states.
        Top row is for policies trained with MLP, bottom row is for policies trained with diffusion policies.
        The x-axis shows the normalised distance to the in-distribution manifold, where the value at 0 represents the in-distribution performance.
        The y-axis shows the mean and standard deviation of final rewards across seeds (higher reward is better).} 
        \label{fig:statistics}
        \vspace{-4mm}
    \end{figure*}

\section{Problem Space Transformations}
\label{sec:problem_space_transformation}
    
In order to enable OOD generalisation over $\mathcal{X^\star}$, we propose to apply a transformation $\mathcal{T}: \mathcal{P \to Q}$, and thus introduce a transformed problem space $\mathcal{Q}$ over which the policy is learned\,\footnote{To ease notation, we will overload $\mathcal{T}$ to also operate over states and actions separately.}.
Let $\hat \pi_\theta$ be the minimiser of the empirical BC loss over the transformed dataset $\mathcal{T(D)} = \{(\mathcal{T}(\mathbf{x}, \mathbf{a}) \mid (\mathbf{x}, \mathbf{a}) \in~\mathcal{D}\}$.
The goal of this transformation is to maximise data coverage over the desired manifold\,\footnote{$\mid\cdot\mid$ represents an $n$-dimensional volume or Lebesgue measure.}: $\min_\mathcal{T} \mid~\mathcal{T(X^\star)} \setminus \mathcal{T(\hat X})\mid$, while ensuring that the BC solution can recover the demonstrator's actions  from the transformed state space: $\pi_d(\mathbf{x}) \approx \mathcal{T}^{-1}(\hat \pi_\theta(\mathcal{T}(\mathbf{x}))) \; \forall \mathbf{x} \in \mathcal{\hat X}$.
We visualise this operation in \Cref{fig:minimisation_visualisation}.
Intuitively, while the objective can be optimised by ``removing information" (e.g. via a low-rank linear projection), the constraint ensures that any task-relevant information is retained.
If the demonstrator $\pi_d$ fulfils certain assumptions, then the problem space may contain irrelevant information, and the objective can be optimised.
It is important to highlight that the just stated objective is introduced as an explanatory device: in this work we don't propose that the optimisation be solved analytically or numerically.
In the following, we discuss practical assumptions which we use to derive transformations which minimise the stated objective, without guarantees that these transformations are strictly optimal. \looseness -1

\subsection{Practical Assumptions for Robotic Manipulation}
This works leverages two assumptions.
First, a well-known property of many manipulation demonstrations is equivariance to transformations in $\mathrm{SE}(n)$ (where $n$ is equal to 2 or 3, depending on the problem dimension) with respect to (w.r.t.) to a fixed world frame $\mathcal{W}$.
For example, in picking up an object with a parallel gripper, what is relevant is the relative location of the gripper to the object, not the absolute location in $\mathcal{W}$.
The second assumption is that object manipulation often affects objects \textit{locally}.
As a consequence, it is often sufficient to have complete information about the surroundings of the EE.
For example, the exact position and orientation of an entity are not decisive when the EE is further away from the object than a distance $\lambda \in \mathbb{R}^+$. 
This distance $\lambda$ is task specific, and excessively low values might make the demonstrator's policy irrecoverable in the transformed problem space.
Nonetheless, for most tasks, we hypothesise that an appropriate value can easily be determined. \looseness -1 %with low effort. 
\subsection{Applying Assumptions in Problem Space}

\paragraph{Transformation $\mathcal{T}_1$}
The first transformation we consider encodes $\mathrm{SE}(n)$ equivariance to changes of the entities' poses w.r.t. to a fixed world frame $\mathcal{W}$, where $\mathcal{W}$ measures the state values in a Cartesian coordinate system with fixed origin, denoted as $W$ (e.g., at the base of the robot arm).
A state $\mathbf{x} \in \mathcal{X}$ can be expressed as  $\mathbf{x} = [\,\prescript{}{\mathcal{W}}{\mathbf{x}_r}, \prescript{}{\mathcal{W}}{\mathbf{x}_{o,1}}, ..., \prescript{}{\mathcal{W}}{\mathbf{x}_{o,N_O}}\,]$, where the prescript denotes the frame in which the state is measured.
Actions are expressed analogously $\mathbf{a} = [\prescript{}{\mathcal{W}}{\mathbf{a}_r}]$.
We propose to transform $\mathcal{X}$ to a frame $\mathcal{E}$ matching the position and the orientation of the end-effector.
% Additionally to the change of the frame of reference, the EE pose can be removed from the state space as it has a constant (zero) value.
This induces a transformed state $\mathcal{T}_1(\mathbf{x}) = [\,\prescript{}{\mathcal{E}}{\mathbf{x}_r}, \prescript{}{\mathcal{E}}{\mathbf{x}_{o,1}}, ..., \prescript{}{\mathcal{E}}{\mathbf{x}_{o,N_O}}\,]$.
Any action\,\footnote{This transformation can be applied no matter if the action is supplied as a next EE position, an offset to the current position or as a velocity, though the exact transformation varies.} $\mathbf{a} \in \mathcal{A}$ is also transformed accordingly: $\mathcal{T}_1(\mathbf{a}) = [\prescript{}{\mathcal{E}}{\mathbf{a}_r}]$.
This ensures that interactions between end-effector and entities may have the same representation, regardless of poses in the fixed coordinate frame $\mathcal{W}$.
In turn, this can increase the density of the occupancy over the transformed state space, effectively enlarging the in-distribution manifold.
An important aside to $\mathcal{T}_1$: For some MDP modelling choices, an additional entity corresponding to a fixed point (usually the target position) in $\mathcal{W}$ needs to be added to $\mathcal{X}$ when transforming the problem space to frame $\mathcal{E}$ such that the demonstrator's policy remains recoverable. 

\paragraph{Transformation $\mathcal{T}_2$}
The second transformation we consider encodes the assumption that manipulation largely occurs locally.
Starting from the output of $\mathcal{T}_1$, we introduce a parameter $\lambda \in \mathbb{R}^+$ and project the position of each entity $\mathbf{x}_{o,i}$ with $i \in [1, \dots, N_O]$ to the surface of the $\lambda$-ball centred in the origin:
\begin{equation}
    \text{proj}(\textbf{x}_{o,i}) = \begin{cases}
        (\text{pos}(\mathbf{x}_{o,i}), \text{rot}(\mathbf{x}_{o,i})), & \text{if } \|\text{pos}(\mathbf{x}_{o, i})\|_2 < \lambda \\
        (\frac{\lambda \text{pos}(\mathbf{x}_{o,i})}{\|\text{pos}(\mathbf{x}_{o,i})\|_2}, \text{rot}(\mathbf{x}_{o,i})), & \text{otherwise}
    \end{cases}
\end{equation}
where $\text{pos}(\cdot)$ and $\text{rot}(\cdot)$ denote the positional and orientational parts of the object pose vectors, respectively.
For $(\mathbf{x}, \mathbf{a}) \in \mathcal{X \times A}$, the transformation $\mathcal{T}_2$ can thus be written as 
\begin{equation}
\begin{split}
    \mathcal{T}_2(\mathbf{x}) &= [\,\text{proj}(\prescript{}{\mathcal{E}}{\mathbf{x}_r}),\text{proj}(\prescript{}{\mathcal{E}}{\mathbf{x}_{o,1}}), ..., \text{proj}(\prescript{}{\mathcal{E}}{\mathbf{x}_{o,N_O}})\,],
    \\
    \mathcal{T}_2(\mathbf{a}) &= [\prescript{}{\mathcal{E}}{\mathbf{a}_{r}}].
\end{split}
\label{eq:T2}
\end{equation}
Small values of $\lambda$ can greatly reduce the size of the transformed desired manifold $\mathcal{T}_2(\mathcal{X}^\star)$ (i.e., by ``clipping" it), and thus maximise data coverage over it.
As long as the demonstrator is concerned with local information (i.e., it is invariant to the distance of entities that are further away than $\lambda$), its policy can be recovered after the transformation.
On the other hand, if $\lambda$ is too small, an arbitrary policy might be irrecoverable as $\mathcal{T}_2$ effectively loses information (i.e., on the exact position of distant entities).
% The significance of $\prescript{}{\mathcal{E}}{\mathbf{x}_r}$ not being projected is discussed below.

\paragraph{Notes on Practical Implementation}
When predicting a single action from a single observation, i.e. when $T_X = T_A = 1$, in both $\mathcal{T}_1$ and $\mathcal{T}_2$ the pose of the end-effector in frame $\mathcal{E}$, $\prescript{}{\mathcal{E}}{\mathbf{x}_r}$, can be dropped, as it is a zero vector and identity orientation.
When using a method to predict an action trajectory from a trajectory of states, the pose of the end-effector is retained, as not all steps will contain trivial information. 
% Care needs to be taken for selecting the which ee pose is used to determine E, and while we don't think that there is much difference, in our experiments we use the most current EE pose in the trajectory of observations.
Slight care needs to be taken for selecting which end-effector pose is used to determine frame $\mathcal{E}$.
We propose that the pose of the most current time step is used.
%The end-effector pose doesn't need to be projected as the assumption behind transformation $\mathcal{T}_2$ is that manipulation occurs locally.
% A natural extension is that in the state history the end-effector has travelled locally and furthermore current methods don't use long state horizons.
While we propose that the end-effector pose $\prescript{}{\mathcal{E}}{\mathbf{x}_r}$ is projected in \cref{eq:T2}, as in principle this follows from the assumption, just like the projection of the other state entries, in practice the state horizons $T_X$ are short, and thus the projection would not have an effect.
In the experiments (see \cref{sec:result}) this is the case as well, thus we cannot provide any experimental data on the effect of the projection on $\prescript{}{\mathcal{E}}{\mathbf{x}_r}$. 
% The EE pose doesn't need to be projected as the transformation assumes that manipulation occurs locally and it is reasonable to assume that the end-effector has also only travelled locally. 
% Especially as current methods don't use long horizons on the observation history this is an additional reason.

\section{Experimental Results}
\label{sec:result}
    \begin{figure*}[t]
        \centering
        \vspace{-6mm}\includegraphics[width=0.75\textwidth]{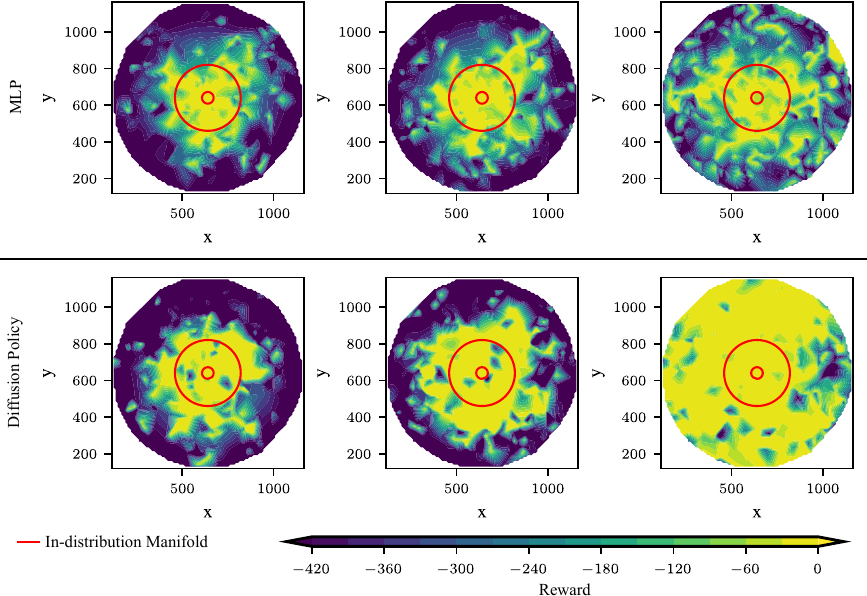}
        \vspace{-3mm}
        \centering
        \caption{Comparison of Baseline $\mathcal{P}$ (left), $\mathcal{T}_1(\mathcal{P})$ (middle) and $\mathcal{T}_2(\mathcal{P})$ (right) for \texttt{PushT}. Plot colour indicates (interpolated) reward per initial object position averaged over seeds. 
        The top row shows the results for an MLP, the bottom row for diffusion policies.
        The in-distribution manifold lies within the red torus (between the inner and outer circle), the OOD manifold outside of the outer red circle.
        The plot in all cases visualises the baseline problem space $\mathcal{P}$. \looseness -1}
        \label{fig:pusht_2d}
        \vspace{-6mm}
    \end{figure*}

    In this section we want to understand how BC policies perform in-distribution and OOD, in the baseline problem space $\mathcal{P}$, and in those defined by the proposed transformations. We first consider three simulated tasks which fulfil our assumptions on the problem-type to evaluate the effect of the transformations on OOD generalisation.
    While we focus on simulated experiments as they allow better control over experimental conditions, we complete our empirical evaluation on a real-world task in Section \ref{sec:real}, which confirms simulated results on hardware. \looseness -1

        \begin{figure}[b]
        \centering
        \vspace{-3mm}
        \includegraphics[width=\linewidth]{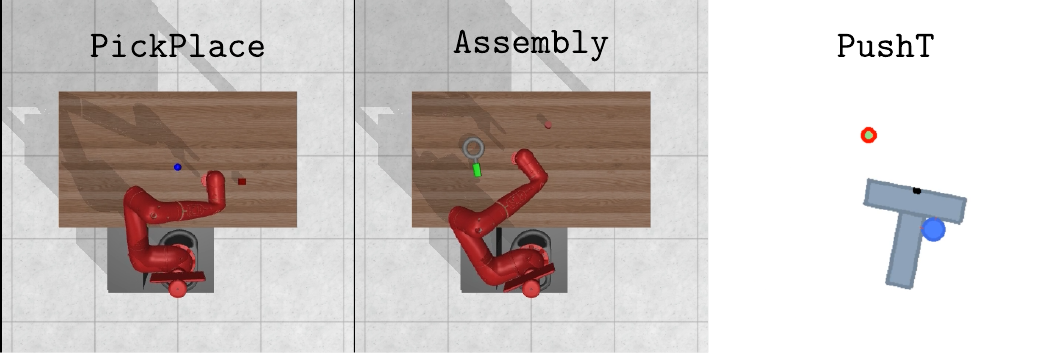}
        \vspace{-3mm}
        \caption{A rendering of the three simulation environments considered in our evaluation.}
        \label{fig:task_vis}
        \vspace{-3mm}
    \end{figure}

    \subsection{Evaluation Tasks}
    \paragraph{Meta-World Tasks}
    In \texttt{PickPlace} and \texttt{Assembly}, the demonstrator controls the position of the end-effector of a 7-DOF robot arm in a 3D environment.
    Both tasks are adapted from the Meta-World benchmark~\cite{yu2019meta}.
    
    In \texttt{PickPlace}, the task is to move a block from an arbitrary initial position to a goal position fixed above the table.
    We modify the task by replacing the object from the original cylinder to a cube.
    The policy only controls the end-effector position for this task, the commanded end-effector orientation is fixed (this is as is given in the benchmark).
    The demonstration data is collected using the scripted policy provided by the benchmark.
    
    In \texttt{Assembly}, the task is to pick up a tool with a handle and a loop and then to place the tool with the loop around the peg.
    We modify the environment compared to benchmark to make the task more realistic.
    The position and orientation of the tool are randomised at initialisation.
    The policy controls the 3D position of the end-effector and a subspace of the orientation.
    While the end-effector orientation command given to the environment is fixed such that it stays parallel to the table, the policy commands the rotation angle $\alpha$ in that plane.
    The demonstration data is provided by a scripted policy, which was modified based on the one provided by the benchmark.
    \paragraph{PushT Environment}
    \texttt{PushT} is a simulated 2D environment, where the demonstrator controls a 2D point mass end-effector, adapted from~\cite{chi2023diffusionpolicy}.
    We have modified the environment compared to the version presented in~\cite{chi2023diffusionpolicy}.
    The task is to move the T from its initial pose to the goal position at the centre of the environment. 
    A constant point on the T needs to reach the target - unlike the goal in~\cite{chi2023diffusionpolicy}, the final pose can be arbitrary.
    The demonstration data is collected by a human demonstrator.
    A visualisation of all three task environments is shown in \Cref{fig:task_vis}.

    \subsection{Evaluation Methodology}
    We test the effectiveness of our proposed problem space transformations when learning a BC policy with two different BC methods. 
    The first method is the naive approach to train a policy which maps a single state to a single action by minimising the mean-squared error of the policy predictions using MLPs.
    The second method is to predict a trajectory of actions from a trajectory of states using a transformer-based diffusion policy approach, as proposed in \cite{chi2023diffusionpolicy}.
    The basic experimental setup is the same for all environments.
    A manifold of the state space is defined and used to sample initial states.
    Demonstrations are collected starting from these sampled initial distributions.
    At test time the learned policies are tested by initialising the environments at a fixed set of initial states.
    To measure in-distribution performance these initial states are sampled in the same region as that used in data collection.
    OOD performance is measured by sampling initial states which lie in a region which doesn't intersect with the data collection manifold.
    We train policies with both policy classes on the baseline problem space $\mathcal{P}$, on the problem space transformed into the end-effector space $\mathcal{T}_1(\mathcal{P})$ and on the problem space additionally transformed using the projection $\mathcal{T}_2(\mathcal{P})$.
    The same model hyperparameters per task and policy representation are used per task to ensure correct comparison.
    Implementation details for the learning algorithms and hyperparameters can be found in \Cref{sec:implementation_details}.
    Details on the datasets and the in-distribution and OOD manifold specifications are in \Cref{sec:task_details}. \looseness -1

    \subsection{Main Results}
    
    \Cref{fig:statistics} reports the final rewards for all tasks and problem spaces, as a function of the distance of the initial configuration with respect to those for which data was collected.
    From left to right, the initial position of entities (e.g. T, cube or ring with handle) is sorted into bins (for evaluation purposes only), which expand concentrically from the in-distribution manifold.
    From left to right per plot we report the mean reward per initial state bin with the normalised distance from the in-distribution manifold reported on the x-axis.
    The value at distance zero represents the in-distribution performance while the others represent OOD bins with increasing distance. 
    The distances are reported on a normalised axis, though they are not equal for all tasks.
    We report the results both for an MLP policy and Diffusion Policy.
    For both policy types, the in-distribution validation performance is similar in all three problem spaces.
    For both policy types all problem spaces display some performance degradation moving from in-distribution to the maximum OOD bin.
    In certain cases $\mathcal{T}_1$ is able to aid OOD generalisation.
    At the same time, $\mathcal{T}_2$ performs at least equally as well as either other problem space and in some cases significantly better, which highlights the relative importance of locality to $\mathrm{SE}(n)$ equivariance.
     
    In \Cref{fig:pusht_2d}, the heat map of final rewards per initial object (T) position in \texttt{PushT} is shown, providing a detailed comparison between the three problem spaces for both policy classes.
    The plots always visualise the baseline problem space $\mathcal{P}$.
    In general it can be seen that Diffusion Policy is better able to learn the task than the naive MLP policy representation, already for the in-distribution region.
    % The policy in $\mathcal{P}$ is able to solve the task for initial states of the object which lie on the fringes of the in-distribution manifold, and not far beyond. 
    While the policy is able to successfully complete the task for initial states of the object which lie on the fringes of the in-distribution manifold in $\mathcal{P}$, it generally fails not far beyond.
    The performance improves for $\mathcal{T}_1$, while
    $\mathcal{T}_2$ is able to successfully execute tasks far outside the in-distribution manifold.
    This is especially visible for BC trained with Diffusion Policy, where in the transformed problem space $\mathcal{T}_2(\mathcal{P})$ high reward is achieved for most initial states - in-distribution as well as OOD. \looseness -1
    Heat maps for \texttt{PickPlace} and \texttt{Assembly} can be found in \Cref{sec:add_results}.

    \begin{figure}
        \centering
        \includegraphics[width=0.8\linewidth]{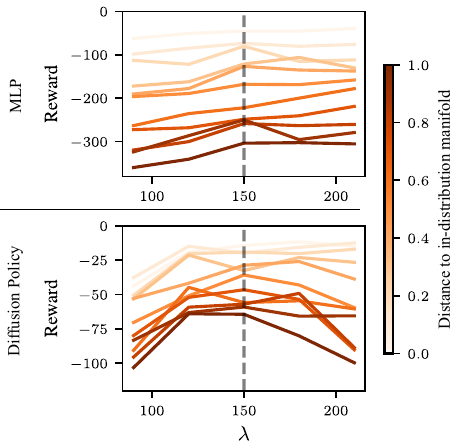}
        \vspace{-4mm}
        \caption{Ablation of $\lambda$ (in pixels) for \texttt{PushT} on in-distribution and OOD performance. 
        BC with MLP above and Diffusion Policy below.
        The dotted line represents the value selected for the experiments.}
        \vspace{-6mm}
        \label{fig:lambd_ablation}
    \end{figure}

    \subsection{Ablations}
    The main hyperparameter introduced by the proposed method is the projection range $\lambda$. 
    It must be large enough to retain enough information about the local manipulation task, while keeping it as small as possible to boost OOD performance.
    Of course, a $\lambda$ value tending towards infinity means that $\mathcal{T}_2$ loses its effect, and the performance will match $\mathcal{T}_1$.
    In our experiments, $\lambda$ had to generally be just large enough such that the geometry of the object and its interaction with target position doesn't produce values greater than the projection radius.
    At the same time the radius should be kept small enough that it ideally causes in-distribution states to be projected to the $\lambda$-ball, and thus cover it.
    For both policy classes, we run an ablation over a range of $\lambda$ values with three random seeds per value.
    For diffusion policies it can clearly be seen that small $\lambda$ values hurt in-distribution performance (and hence also OOD performance), while large $\lambda$ values reduce OOD performance while the in-distribution performance remains constant.
    For the MLP policy representation, the conclusion from the ablation is not as clear, due to the general decrease in performance of the policy. Nevertheless, we can observe that an excessively small $\lambda$ remains detrimental.

    \subsection{Real-World Evaluation}
    \label{sec:real}
    We finally evaluate our approach on a real-world robotic hardware setup to validate the results obtained in simulation.
    Experiments are conducted on a real-world variant of the \texttt{PushT} task, which differs from the simulated environment and poses additional challenges due to sensing noise and actuation errors.
    
\begin{figure}
        \centering
        \includegraphics[width=0.95\linewidth]{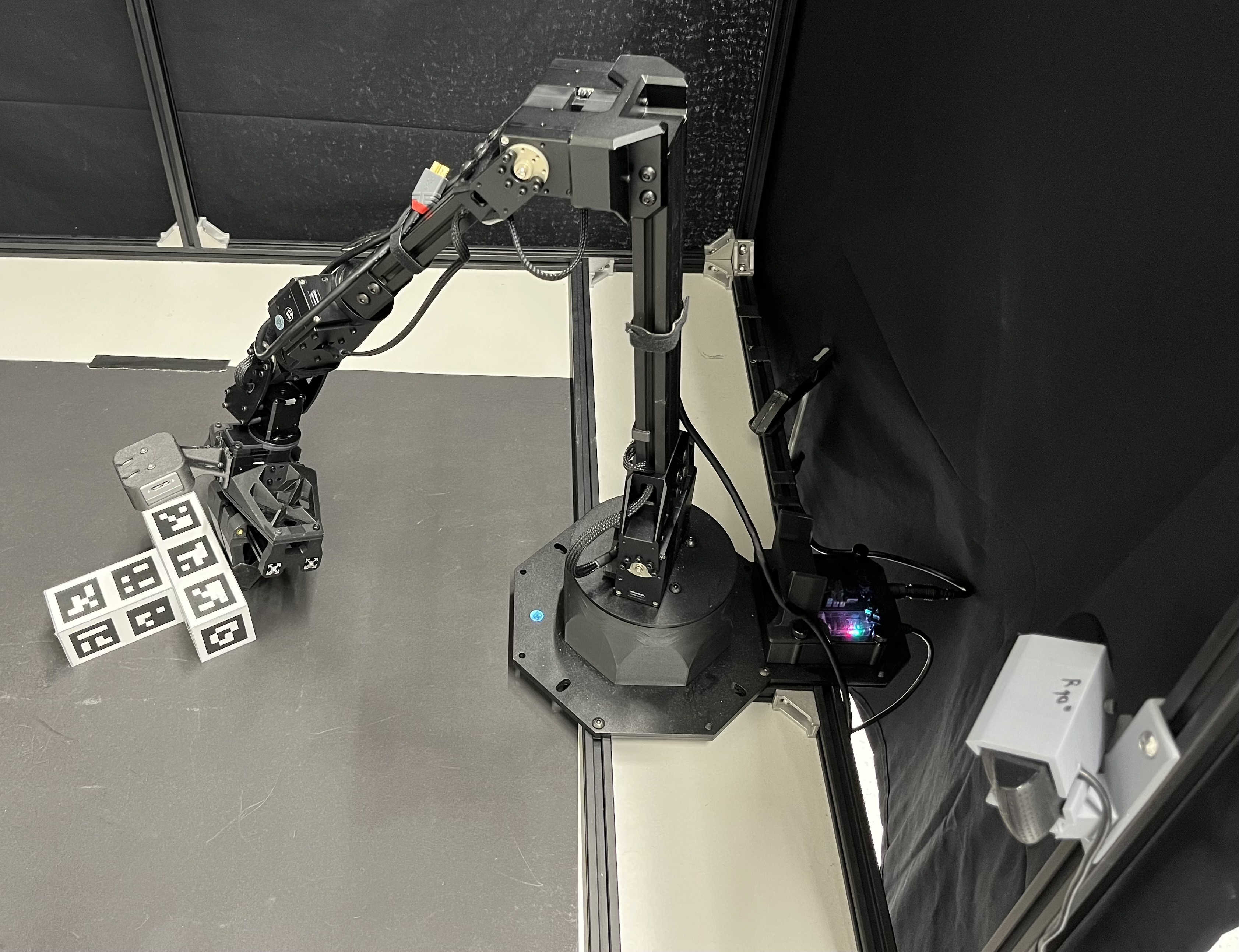}
        \vspace{-2mm}
        \caption{The real-world one-arm setup (using ALOHA \cite{zhao2023learning}) used to collect teleoperation data and to rollout trained policy (wrist camera is unused).}
        \vspace{-6mm}
        \label{fig:real_world_setup}
    \end{figure}

    The task is executed on a single-arm ALOHA robot platform \cite{zhao2023learning}.
    Training data is collected via teleoperation by a single expert demonstrator.
    Consistent with the simulation experiments, the policy receives state-based observations as input.
    The pose of the T-shaped object is estimated using an Apriltag-based tracking system with two calibrated RGB webcams.
    An overview of the hardware setup is shown in \Cref{fig:real_world_setup}.
    
    For the real-world evaluation, we exclusively consider behaviour cloning (BC) with a diffusion policy, due to higher reported performance on hardware \cite{chi2023diffusionpolicy}.
    OOD performance results on the physical robot are summarized in \Cref{tab:real_world_results}.
    For each policy, we execute 15 rollouts from uniformly sampled initial states, drawn from both an in-distribution (ID) set and an out-of-distribution (OOD) set.
    The object is initialised in as similar a location as possible across all compared transformation to ensure fair comparison.
    The first metric we show is the distance from the target the T object should be pushed towards for all three transformations, a lower number is better.
    The second metric shows how much closer to the goal the object is at termination compared to its initial position, as despite best efforts the exact initial state locations varies between rollouts. 
    The third metric represents the same number as the second metric, though it shows the relative distance moved towards the goal.
    
    Consistent with the simulation results, we observe a clear performance improvement when using $\mathcal{T}_2$ compared to the baseline problem space, demonstrating that the benefits of the problem space transformation apply to real-world robotic execution too.
    Though, in contrast with simulation results, we see that the performance of $\mathcal{T}_1$ is reduced compared to the baseline, which in turn means that the increase in performance of $\mathcal{T}_2$ can be traced back to the assumption of locality.

    \begin{table}[h]
    \caption{Real-World Evaluation}
    \centering
    \resizebox{\linewidth}{!}{
    \begin{tabular}{l|ccc}
        \toprule
        Policy & Baseline & $\mathcal{T}_1$ & $\mathcal{T}_2$\\
        \midrule
        Mean dist. at termination [m] & 0.1977 & 0.2075 & 0.1853\\
        Mean dist. moved towards goal [m] & 0.0387 & 0.0292 & 0.0501\\
        Mean rel. dist. moved towards goal [-] & 16.38\% & 12.32\% & 21.29\%\\
        \bottomrule
    \end{tabular}
    }
    
    \label{tab:real_world_results}
\end{table}

\section{Limitations and Conclusion}
\label{sec:conclusion}
    This work validates the hypothesis that the problem space in which a BC policy is trained has an effect on OOD performance. Moreover, it demonstrates that, by choosing appropriate transformations leveraging practical assumptions of common manipulation problems, broader OOD generalisation can be enabled.
    Nevertheless, our experimental validation is restricted to tasks with rigid bodies and low-dimensional state information.
    For instance, the proposed transformations wouldn't be able to directly handle visual input, nor OOD scenarios that would arise in visual data.
    While this domain lies beyond the scope of this work, we still expect that domain-specific properties might be leveraged to design different problem space transformations.
    Finally, the reliance on the fact that the data covers the transformed problem space well remains.
    For these reasons, we suggest that future directions of work could focus on determining transformations which are applicable more generally, or data-driven methods which aid in the discovery of transformations. \looseness -1


%%%%%%%%%%%%%%%%%%%%%%%%%%%%%%%%%%%%%%%%%%%%%%%%%%%%%%%%%%%%%%%%%%%%%%%%%%%%%%%%

%%%%%%%%%%%%%%%%%%%%%%%%%%%%%%%%%%%%%%%%%%%%%%%%%%%%%%%%%%%%%%%%%%%%%%%%%%%%%%%%

%%%%%%%%%%%%%%%%%%%%%%%%%%%%%%%%%%%%%%%%%%%%%%%%%%%%%%%%%%%%%%%%%%%%%%%%%%%%%%%%
\section*{APPENDIX}

\subsection{Implementation Details}
\label{sec:implementation_details}
\paragraph{BC with MLPs}
We train the MLP models using PyTorch.
The policy is implemented as a deterministic MLP with ReLU activation functions.
We use dropout and $L_2$ regularisaition.
The models are trained using mini random batches and all data is standardised to zero mean and one standard deviation before being passed to the neural network.
The loss function is the 2-norm between policy prediction and dataset sample.
The Adam optimizer is used to update the weights.
Training time for all tasks is in the order of minutes, while evaluation in simulation is on the order of 10 minutes.
The hyperparameters used in training the models is shown in \Cref{tab:mlp_params}.

\begin{table}[h]
    \caption{Training hyperparameters used when training MLP BC policy for the different tasks.}
    \centering
    \begin{tabular}{l|ccc}
        \toprule
        Task & \texttt{PickPlace} & \texttt{Assembly} & \texttt{PushT}\\
        \midrule
        Hidden Layers & 5 & 5 & 5\\
        Hidden Dimen. & 512 & 512 & 512\\
        Dropout prob. & 0.05 & 0.05 & 0.05\\
        Regularisation weight & 1e-5 & 1e-5 & 1e-5\\
        Learning Rate  & 1e-3 & 1e-3 & 1e-3 \\
        Batch Size & 512 & 512 & 1024\\
        Epochs& 600 & 600 & 1200\\
        \bottomrule
    \end{tabular}
    
    \label{tab:mlp_params}
\end{table}

\paragraph{BC with Diffusion Policy}
We use a PyTorch implementation of Diffusion Policy based on the openly available implementation provided by the authors.
We follow the same approach to training the models as described in \cite{chi2023diffusionpolicy} using the transformer-based variant.
Training time for all tasks with our implementation is on the order of 10 hours, though significant speed up would be possible with a more efficient implementation of the transformations.
The hyperparameters used in training are shown in \Cref{tab:diffusion_params}.
The table differentiates between the number of predicted action steps $T_{A_p}$ and the number of action steps which are then executed before policy inference is performed again $T_{A_e}$.
To deploy the diffusion policy on a real system we follow \cite{chi2023diffusionpolicy} by using DDIM noise scheduling and reducing the number of diffusion steps at inference to 16 (100 steps in training) compared with using DDPM noise scheduling in training and testing at 100 diffusion steps for the simulation tasks.
\begin{table}[h]
    \centering
    \caption{Training hyperparameters used when training Diffusion Policy for the different tasks.}
        \label{tab:diffusion_params}
    \begin{tabular}{l|cccc}
        \toprule
        \multirow{2}{*}{Task} & \multirow{2}{*}{\texttt{PickPlace}} & \multirow{2}{*}{\texttt{Assembly}} & \multirow{2}{*}{\texttt{PushT}} & \texttt{PushT}\\
         & & & & \texttt{Real} \\
        \midrule
        $T_X$ & 2 & 2 & 2 & 2\\
        $T_{A_p}$ & 10 & 10 & 16 & 16\\
        $T_{A_e}$ & 8 & 8 & 8 & 6\\
        Layers & 8 & 8 & 8 & 8\\
        Learning Rate  & 1e-4 & 1e-4  & 1e-4 & 1e-4 \\
        Batch Size & 256 & 256 & 256 & 256\\
        Epochs& 5010 & 1800 & 5010 & 1750\\
        Attn. Dropout & 0.3 & 0.3 & 0.01 & 0.1\\
        \bottomrule
    \end{tabular}
    
\end{table}

\subsection{Task and Experiment Details}
\label{sec:task_details}

\begin{table*}[h]
    \centering
    \caption{Summary of task and experiment relevant parameters.}
    \begin{tabular}{l|cccc}
        \toprule
         Task & \texttt{PickPlace} & \texttt{Assembly} & \texttt{PushT} & \texttt{PushT Real}\\
        \midrule
        \# Demonstrations & 100 & 200 & 200 & 200\\
        Collected by & Scripted Policy & Scripted Policy & Human Demonstrator & Human Demonstrator \\
        \# In-dist. evaluation episodes & 100 & 100 & 100 & 15 \\
        \# OOD evaluation episodes & 450 & 450 & 400 & 15 \\
        \multirow{2}{*}{In-dist. manifold obj. pos.} & 
            $x_1 \in [-0.15, 0.15]$m, & $x_1 \in [-0.15, 0.15]$m,  & $r \in [32, 180]$px, & $x_1 \in [0.31, 0.61]$m\\
           & $ x_2 \in [0.48, 0.62]$m  & $ x_2 \in [0.43, 0.67]$m & $\theta \in [0, 2\pi]$rad & $ x_2 \in [-0.15, 0.15]$m \\   
        
        \multirow{2}{*}{OOD manifold obj. pos.} & 
            $x_1 \in [-0.465, 0.465]$m, & $x_1 \in [-0.465, 0.465]$m,  & $r \in [180, 534]$px, & $x_1 \in [0.2, 0.71]$m\\
           & $x_2 \in [0.4, 0.7]$m  & $ x_2 \in [0.38, 0.72]$m & $\theta \in [0, 2\pi]$rad & $ x_2 \in [-0.225, 0.225]$m \\  
        $\lambda$ (for $\mathcal{T}_2$) & 0.1m & 0.2m & 150px & 0.225m\\
        \# Experiment Seeds MLP & 8 & 8 & 8 & -\\
        \# Experiment Seeds DP & 5 & 5 & 5 & 1\\
        Reward & Distance to target & See implementation \cite{yu2019meta} & Distance to target & Distance to target \\
        Action Space & \begin{minipage}[t]{0.25\columnwidth}
             \centering
             [EE Pos., Gripper Force]
        \end{minipage} & 
        \begin{minipage}[t]{0.4\columnwidth}
             \centering
             [EE Pos., 1D EE Orientation, Gripper Force]
        \end{minipage} & [EE Pos. (2D)] & \begin{minipage}[t]{0.4\columnwidth}
             \centering
             [EE Pos., EE Orient. (quater- nion), Norm. Gripper Pos.]
        \end{minipage}  \\
        \bottomrule
    \end{tabular}
    \label{tab:dataset_experiment_details}
\end{table*}

The main task and experiment specific parameters are summarised in \Cref{tab:dataset_experiment_details}. The specified in-distribution and OOD manifolds relate to the position of the relevant object of the task. 
For \texttt{PickPlace}, \texttt{Assembly} and \texttt{PushT Real} the in-distribution manifold is a rectangle and the OOD manifold is a larger rectangle excluding the in-distribution manifold. The parameters of these rectangles are provided in Cartesian coordinates in the table.
For \texttt{PushT}, they are sampled from a torus with inner and outer radius in both cases, the parameters in the table are provided in cylindrical coordinates.
In all cases the end-effector is initialised at the same pose. 
For $\texttt{Assembly}$, $\texttt{PushT}$ and \texttt{PushT Real} the orientation of the relevant object is randomised in $[0, 2\pi]$ in training and evaluation, but for $\texttt{PickPlace}$ the initial orientation of the object is constant.

\subsection{Additional Results}
\label{sec:add_results}
\begin{figure*}
    \centering
    \includegraphics[width=0.95\linewidth]{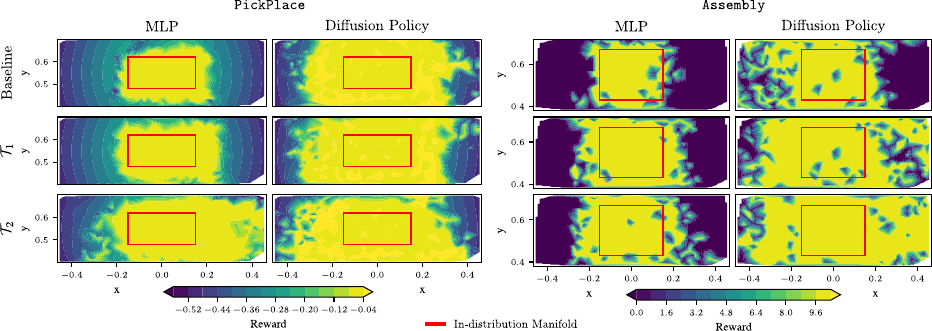}
    \vspace{-2.5mm}
    \label{fig:other_2d_heatmaps}
    \caption{\looseness -1 Heat map of results for \texttt{PickPlace} on the left and \texttt{Assembly} on the right for all three problem spaces. Plot colour indicates (interpolated) reward per initial object position. The in-distribution manifold lies within the red rectangle, the OOD manifold outside of it for both tasks. The plots in all cases visualise the baseline problem space $\mathcal{P}$.}
\end{figure*}

\Cref{fig:other_2d_heatmaps} shows the heat map results for \texttt{PickPlace} and \texttt{Assembly}. \looseness -1

\section*{ACKNOWLEDGMENT}

K.D. extends his gratitude to Núria Armengol, Yijiang Huang and Miguel Zamora for valuable and insightful discussions. M.B. is supported by the Max Planck ETH Center for Learning Systems.
\addtolength{\textheight}{-10.5cm}

%%%%%%%%%%%%%%%%%%%%%%%%%%%%%%%%%%%%%%%%%%%%%%%%%%%%%%%%%%%%%%%%%%%%%%%%%%%%%%%%

% References are important to the reader; therefore, each citation must be complete and correct. If at all possible, references should be commonly available publications.

\bibliographystyle{IEEEtran}
\bibliography{references.bib}

\end{document}